\pdfoutput=1
\documentclass[sigconf,screen,nonacm]{acmart}

\usepackage[ruled,vlined]{algorithm2e}
\usepackage{subcaption}
\usepackage{microtype}

\AtBeginDocument{%
  }

\begin{document}

\title{FlashMoE: Reducing SSD I/O Bottlenecks via ML-Based Cache Replacement for Mixture-of-Experts Inference on Edge Devices}


\author{Byeongju Kim}
\affiliation{%
  \institution{KAIST}
  \country{Republic of Korea}
}
\email{byeongju.kim@kaist.ac.kr}

\author{Jungwan Lee}
\affiliation{%
  \institution{KAIST}
  \country{Republic of Korea}
}
\email{milelion@kaist.ac.kr}

\author{Donghyeon Han}
\affiliation{%
 \institution{Chung-Ang University}
  \country{Republic of Korea}
}
\email{hdh4797@cau.ac.kr}

\author{Hoi-Jun Yoo}
\affiliation{%
  \institution{KAIST}
  \country{Republic of Korea}
}
\email{hjyoo@kaist.ac.kr}

\author{Sangyeob Kim}
\affiliation{%
  \institution{Yonsei University}
  \country{Republic of Korea}
}
\email{sangyeob.kim@yonsei.ac.kr}


\begin{abstract}
  Recently, Mixture-of-Experts (MoE) models have gained attention for efficiently scaling large language models. Although these models are extremely large, their sparse activation enables inference to be performed by accessing only a fraction of the model at a time. This property opens the possibility of on-device inference of MoE, which was previously considered infeasible for such large models. Consequently, various systems have been proposed to leverage this sparsity and enable efficient MoE inference for edge devices. However, previous MoE inference systems like Fiddler\cite{fiddler} or DAOP\cite{daop} rely on DRAM-based offloading and are not suitable for memory-constrained on-device environments. As recent MoE models grow to hundreds of gigabytes, RAM-offloading solutions become impractical. To address this, we propose \textbf{FlashMoE}, a system that offloads inactive experts to SSD, enabling efficient MoE inference under limited RAM. FlashMoE incorporates a lightweight ML-based caching strategy that adaptively combines recency and frequency signals to maximize expert reuse, significantly reducing storage I/O. In addition, we built a user-grade desktop platform to demonstrate the practicality of FlashMoE. On this real hardware setup, FlashMoE improves cache hit rate by up to 51\% over well-known offloading policies such as LRU and LFU, and achieves up to 2.6× speedup compared to existing MoE inference systems.
\end{abstract}



\keywords{Mixture-of-Experts, Offloading, LLM Inference, Scheduling, Cache Replacement Policies}


\maketitle

\section{INTRODUCTION}
Large language models (LLMs) have continued to grow rapidly in size, driven by consistent improvements in performance. This trend is largely explained by the scaling law\cite{scalinglaw}, which show that model performance increases predictably with more parameters, more data, and more computation. As a result, state-of-the-art LLMs are being trained with tens or even hundreds of billions of parameters. However, as model size increases, so do the demands on memory and computation for both training and inference. This makes scaling a central challenge in modern LLM development.

\begin{figure}
  \centering
  \includegraphics[width=1\linewidth]{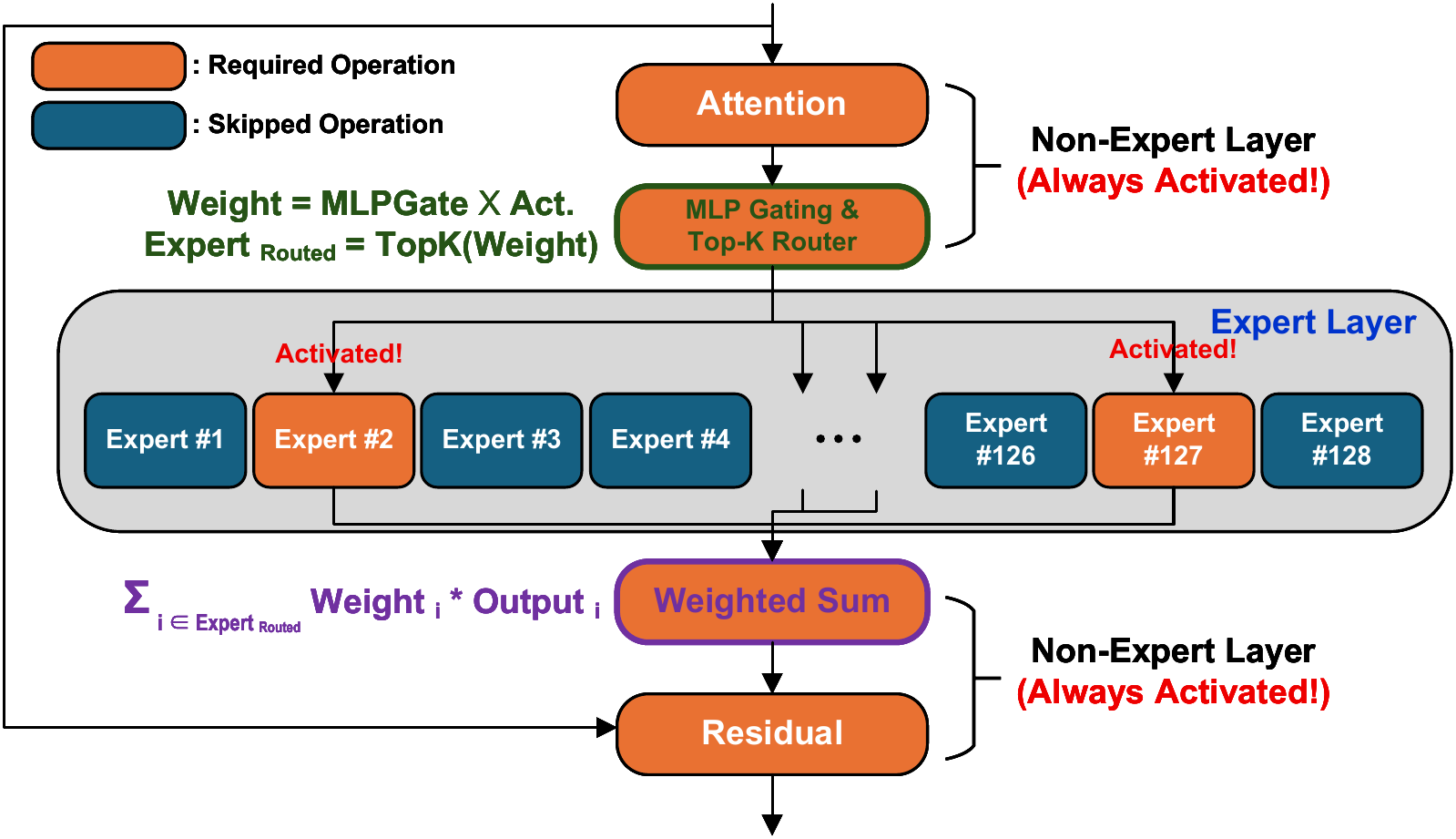} 
  \caption{Illustration of an MoE model decoder layer.
}
  \label{fig:moelayer}
\end{figure}

To address this, the Mixture-of-Experts (MoE)\cite{sparsemoe} architecture has emerged as a key strategy. They achieve low computational cost through sparse activation, where only a small subset of expert networks is used at each forward pass. This design allows models to scale to hundreds of billions of parameters without linearly increasing training or inference cost. For example, Qwen3-30B-A3B\cite{qwen3} has 3.3B sparsely activated parameters, while having total 30.3B parameters. Consequently, MoE has been widely adopted in recent high-performance LLMs such as Gemini\cite{gemini}, and DeepSeek\cite{deepseek}. By decoupling model capacity from compute cost, MoE provides a practical and efficient path to pushing LLMs further.

If inactive experts can be offloaded and only necessary subsets are loaded on demand, MoE enables large LLMs to run on edge devices with reduced memory requirements. Existing offloading-based frameworks such as Fiddler~\cite{fiddler}, DAOP~\cite{daop}, and Pre-gated MoE~\cite{pregatedmoe} implement expert caches but are not optimized for on-device deployment. 

To address this, we propose \textbf{FlashMoE}, a system-level framework for efficient MoE inference:

\begin{itemize}
  \item \textbf{Software-Level Design:} Experts and non-experts are stored separately, with experts further subdivided at layer and unit levels, enabling fine-grained on-demand loading from SSD.
  \item \textbf{ML-Based Cache Design:} Using ML to approximate Belady’s optimal policy, FlashMoE achieves higher hit rates than heuristic caches, reducing memory footprint and inference latency.
\end{itemize}

\begin{figure*}[!t]
  \centering
  \includegraphics[width=0.85\linewidth]{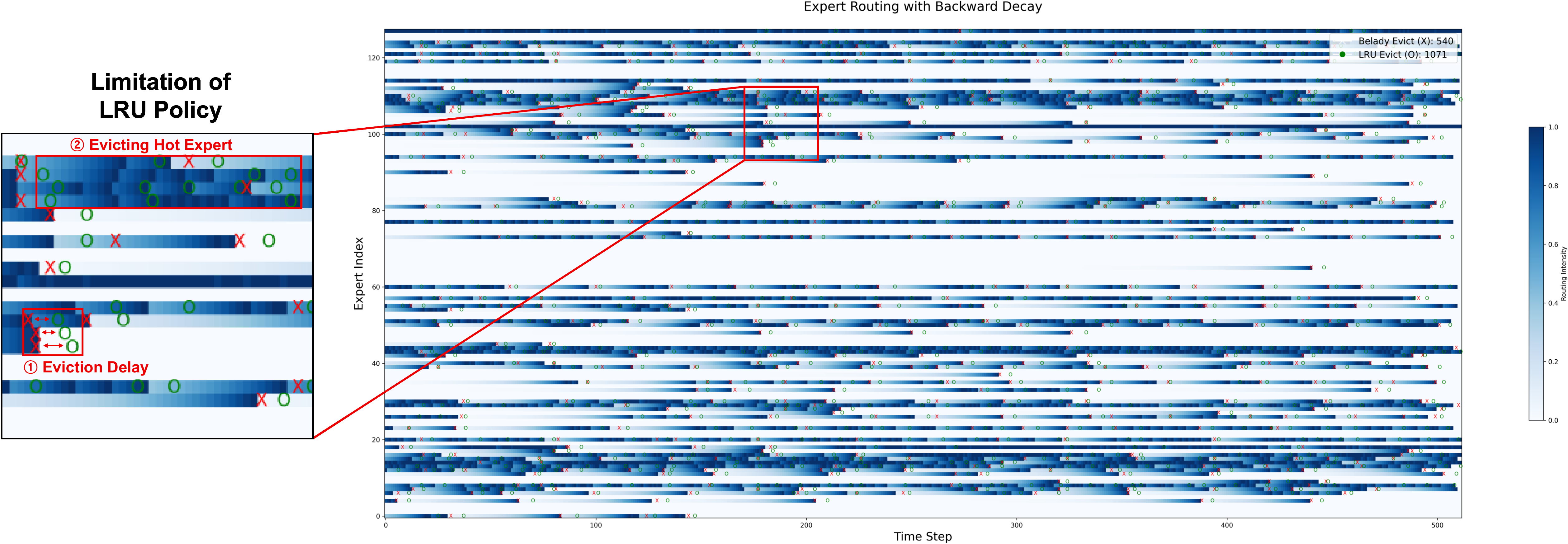} 
  \caption{Heatmap comparing LRU and Belady cache policies. The deepest color means the experts are routed at that time step, and the color changes lighter as time goes backward. Evictions based on Belady policy is marked as red 'X' shape, while LRU is marked as green 'O'. Due to LRU's recency-based replacement policy, it fails to evict experts immediately after routing as the optimal algorithm does. Additionally, frequently accessed experts are sometimes evicted just before routing, highlighting a key limitation of LRU. This indicates the need for incorporating frequency-based metrics such as LFU and leveraging strategies.
}
  \label{fig:lruheat}
\end{figure*}

\section{BACKGROUND \& MOTIVATION}
\subsection{Mixture of Experts}

As shown in Fig. \ref{fig:moelayer}, MoE layers operate differently from traditional dense MLP layers. Instead of applying the same parameters to every input token, a routing gate first computes scores for a set of expert networks based on each token’s hidden state. Among these, only the top-k experts with the highest scores are selected. Each selected expert processes the token independently, and their outputs are weighted and accumulated to produce the result. This sparse activation allows the model to scale its total parameter count without increasing the computational cost per token, as only a small subset of experts is active at a time.

The challenges of MoE arise from its sparse activation characteristic following the gating mechanism, which results in expert layers being extremely larger than non-expert layers. Expert layers refer to multiple MLP layers (called "experts") in the MoE, and only a few of them are activated for each input. On the other hand, non-expert layers include the attention modules, normalization layers, and the routing gate, which decides which experts to use. These are always active and are much smaller (around 5$\sim$7\%) than the expert layers. 

\subsection{Challenges in Memory Consumption of MoE}

MoE inference systems have emerged with placing experts in diverse memory hierarchy, mainly conducting inference by orchestrating data movement between VRAM and DRAM or between GPU and CPU. While this approach reduces reliance on limited VRAM, it implicitly assumes that the entire model can at least reside in DRAM. 

However, this assumption no longer holds for edge-devices. In practical desktop environments where DRAM is typically limited to 16$\sim$64GB, even storing inactive experts in DRAM becomes infeasible. Additionally, available memory may be further limited by other concurrently running processes. Therefore, an effective MoE inference system must support offloading model weights from a lower memory hierarchy level such as SSD, enabling inference without requiring the full model to be resident in either DRAM or VRAM.

This challenge is also closely related to how MoE models are initialized. Conventional approaches\cite{fiddler,daop, eliseev} load the entire model from a single unified file to last-level cache before starting inference. However, since only a subset of expert layers is activated based on routing decisions, these layers can instead be loaded on-demand from the last-level cache during inference. In on-device environments, since the last-level cache is SSD, minimizing model loading latency is critical.

Therefore, to reflect these structural characteristics, an approach that clearly separates experts and non-experts at the file level is necessary. Non-expert layers such as attention, normalization, and routing gates are essential in every layer and must be loaded into memory during the initialization phase. This allows prioritizing the initialization of non-experts first and loading experts on-demand during inference, significantly reducing the overall model loading time.

\subsection{Cache System of MoE}

The expert cache management methods in existing MoE inference systems \cite{fiddler, eliseev} are based on simple replacement policies which is used for page replacement. Many previous studies use traditional cache replacement algorithms such as LRU or LFU, and actually have hit-rate to some extent. However, when inference involves loading experts from SSDs—which have significantly higher I/O latency compared to DRAM—even a single cache miss can incur a substantial performance penalty.

Moreover, experts exhibit not only temporal locality but also routing-specific reuse patterns, such as repeated selection of certain experts across tokens or sequences. Therefore, to minimize costly SSD access and improve overall efficiency, a more sophisticated eviction policy that captures both temporal and routing-aware locality is required.
\begin{figure*}[!t]
  \centering
  \begin{subfigure}{0.70\textwidth}
   \includegraphics[height=5cm]{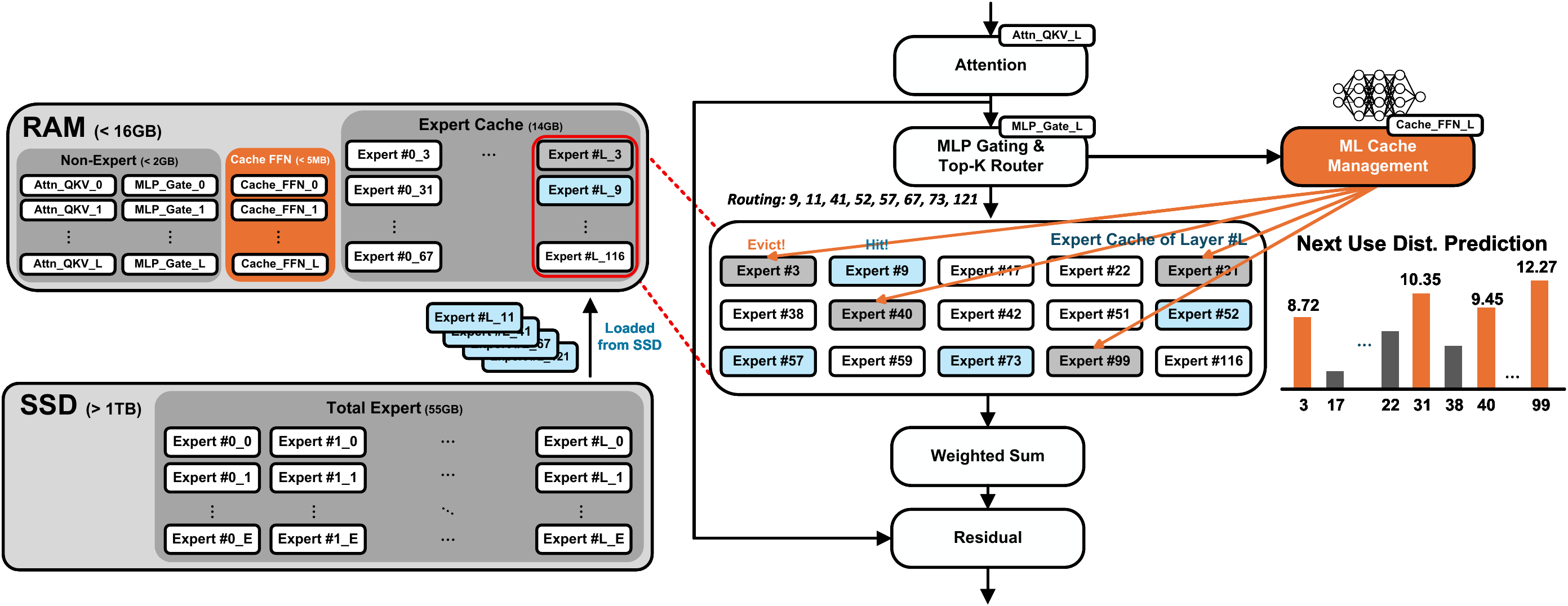}
   \caption{}
  \end{subfigure}\hfill
  \begin{subfigure}{0.28\textwidth}
   \includegraphics[height=5cm]{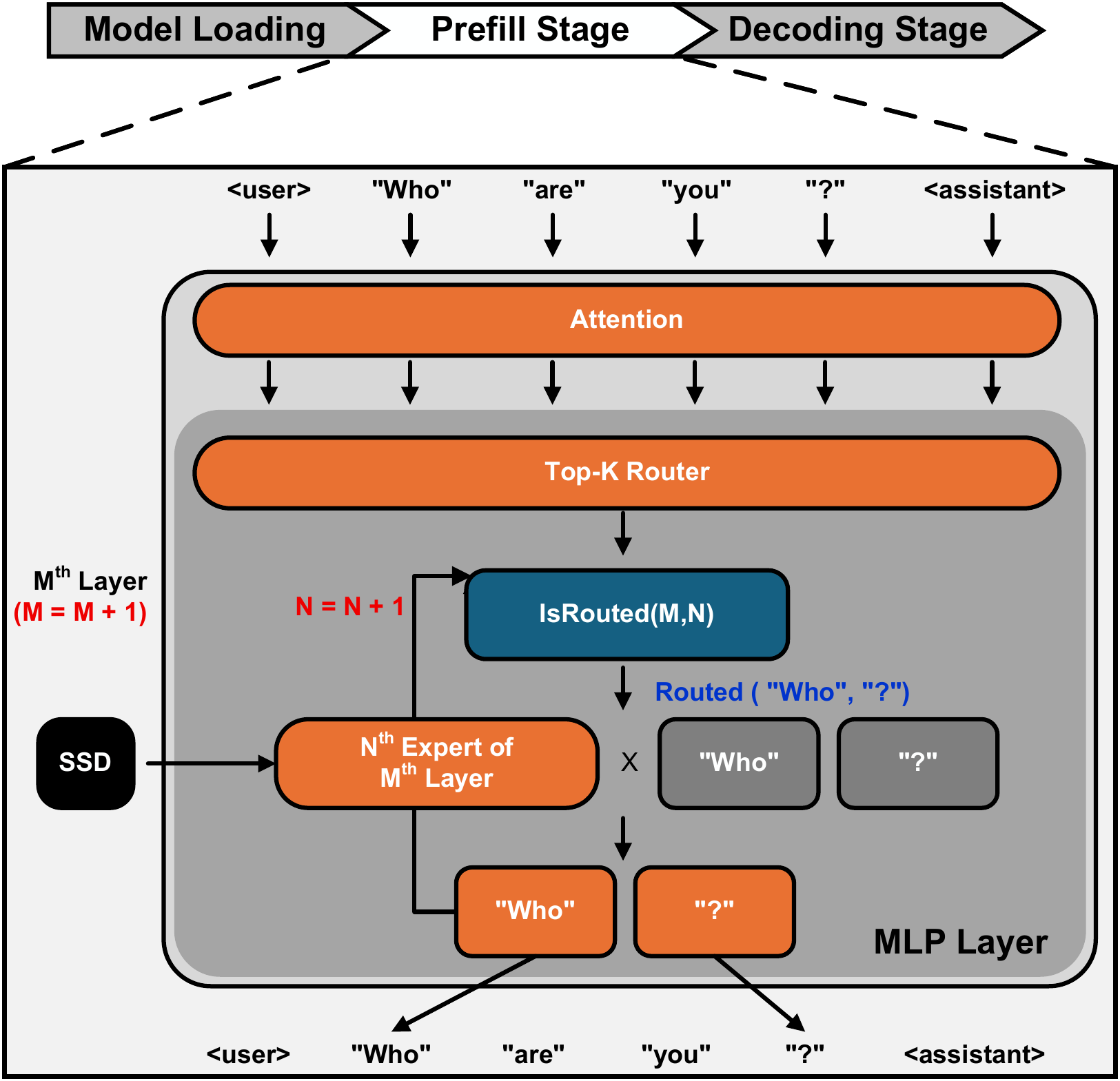}
   \caption{}
  \end{subfigure}\hfill
  \caption{
    (a) Overall system of FlashMoE. 
    (b) FlashMoE's prefill process.
  }
  \label{fig:flashmoe_combined}
\end{figure*}

The experimental results presented in Fig. \ref{fig:lruheat} illustrate the situations that occur when experts are routed according to LRU(\textbf{green 'O'}) and Belady\cite{belady}(\textbf{red 'X'}) replacement policies. Since the Belady policy theoretically provides the optimal replacement timing, immediately replacing a specific expert in such situations is a desirable choice. However, LRU follows a simple recency-based policy, which causes the problem that the expert that should actually be replaced at that moment is not evicted promptly, which we will call it Eviction Delay, as shown in Fig. \ref{fig:lruheat}. Since the experts to be evicted are kept in cache for additional 4 $\sim$ 5 steps, LRU often evicts experts that were recently accessed and are still 'hot', since it does not account for temporal locality beyond recent usage, which we will call it 'Evicting Hot Experts'.

These limitations of LRU (Eviction Delay, Evicting Hot Experts) become more pronounced when an expert that is frequently used but has not been accessed recently is evicted due to recency bias. For example, compared to Belady’s optimal algorithm with an 86\% hit rate, LRU achieves only around 73\%, resulting in nearly 1.9× more I/O operations. These additional misses are often caused by evicting frequently routed experts solely due to a temporary lapse in access, highlighting the inefficiency of LRU in expert-aware caching.

To better illustrate this inefficiency, we measured the re-fetch rate of evicted experts within the following 5 time steps. On average, LRU's evicted experts were reused 34.2\% of the time, which is significantly higher than Belady's 0.1\%. This suggests that LRU tends to evict experts that are still likely to be reused in the near future, revealing substantial room for optimization. Additionally, when comparing eviction decisions between LRU and LFU on the Qwen3-30B-A3B model, we found that LRU made the better choice approximately 56\% of the time. These results highlight that both LRU and LFU policies have room for improvements in cache efficiency for MoE inference systems, and leveraging the two can make a better cache policy targeted for MoE models.
    
As a result, while LRU and LFU are based on their respective criteria (recency and frequency), neither alone sufficiently reflects the characteristics of expert routing. Therefore, it is evident that a more sophisticated, expert-aware cache replacement policy that harmoniously leverages the strengths of both policies is needed.

To tackle these challenges, we propose \textbf{FlashMoE}, a system-level solution tailored for efficient inference of MoE models. Fig. \ref{fig:flashmoe_combined}(a) shows the overall architecture of the proposed FlashMoE system designed to address the previously mentioned issues.

\section{FlashMoE SYSTEM}

\begin{figure*}[!t]
  \centering
  \includegraphics[width=0.85\linewidth]{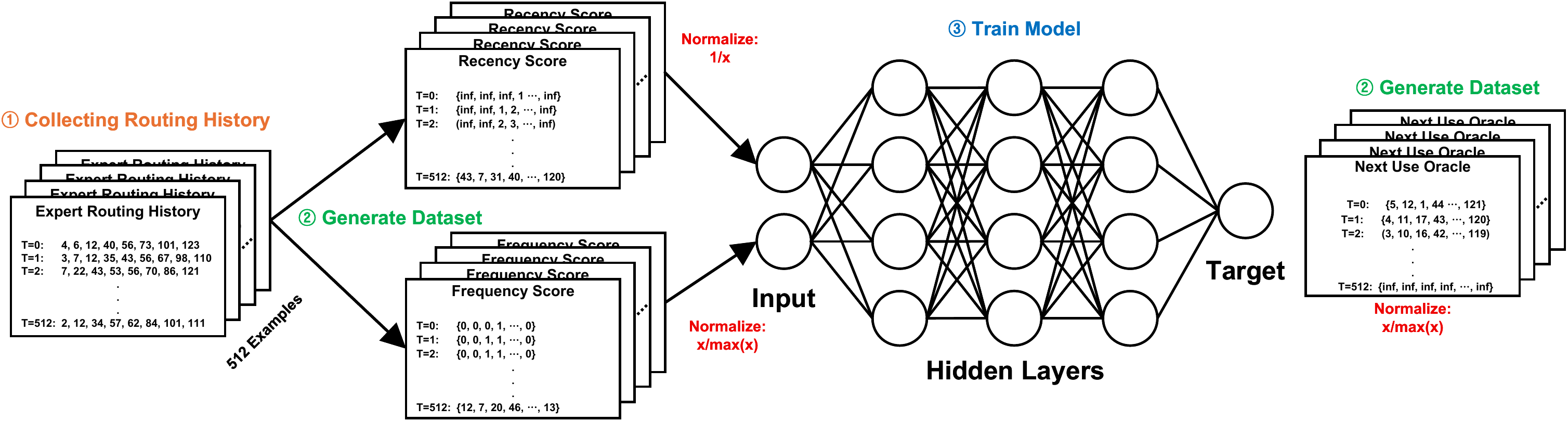} 
  \caption{Overall training method illustration of FlashMoE.}
  \label{fig:training}
\end{figure*}

\begin{figure}[!t]
  \centering
  \includegraphics[width=0.99\linewidth]{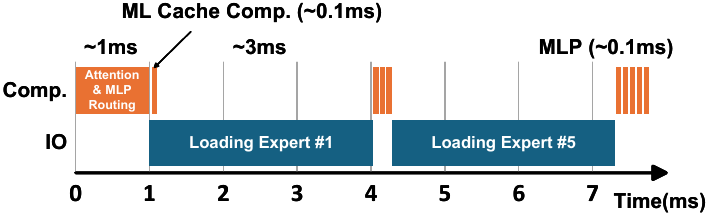} 
  \caption{Decoder layer computation pipeline of FlashMoE.
}
  \label{fig:pipeline}
\end{figure}

\subsection{Overall System}

\textbf{• Model Decomposition \& Loading Stage:} FlashMoE stores MoE model weights by separating experts and non-experts, with experts further divided and saved as individual .pt files (PyTorch model format) according to each layer and expert index. Each expert layer is saved along with its model class and operations, allowing the weights to be loaded directly using the torch.load function without further initialization in the main code. 

For the non-expert components, MLP computations are bypassed by setting hidden states to zero in the overridden forward function, effectively zeroing out these layers to reduce model size and accelerate initialization. As a result, the non-expert model retains minimal parameters and incurs negligible computational overhead. At inference time, FlashMoE initially loads only the non-expert modules into memory, significantly reducing the loading latency compared to full-model loading.

\textbf{• Prefill Stage:}  As shown in Fig. \ref{fig:flashmoe_combined}(b), when tokens are routed to experts, FlashMoE identifies which experts are accessed across the entire token batch. It then loads each required expert file on-demand from the SSD exactly once per inference iteration. The expert-specific MLP computations are executed, and their outputs are redistributed and accumulated back to corresponding token indices through indexed addition. This mechanism preserves exact attention calculations while minimizing expert loading operations and memory usage.

Routed expert weights are loaded from storage and their MLP computations are performed on one or multiple tokens simultaneously, as illustraed in Fig. \ref{fig:flashmoe_combined}(b). Since the cache size is limited, FlashMoE must decide which experts to retain in the cache and which to evict, which means that the experts are freed. This decision-making constitutes the offloading strategy of the prefill stage.

\textbf{• Decoding Stage:} FlashMoE implements a configurable cache system that manages expert weights on a per-layer basis by allocating a fixed number of experts allowed to reside in VRAM for each layer. By adjusting this cache size, users can control the trade-off between VRAM usage and decoding speed, scaling resource consumption according to the available hardware capacity.

Given the available VRAM size, the maximum cache size per layer can be approximated as follows:
\vspace{0.0em}
{\small
\begin{equation}
\text{$cache$ $size$}
= \bigl(\text{$VRAM$ $size$} - size_{\text{non-expert}}\bigr) \times \frac{total\_experts}{size_{\text{expert}}}
\end{equation}
}
where
\begin{itemize}
    \item $size_{\text{non-expert}}$ denotes the memory used by non-expert components,
    \item $size_{\text{expert}}$ is the memory required to store all experts,
    \item $total\_experts$ is the total number of experts\ per layer.
\end{itemize}

When decoding starts under a limited cache size, experts must be loaded into VRAM, and existing experts are evicted from the cache as needed. As shown in Fig.~\ref{fig:pipeline}, expert loading accounts for over 70\% of the total decoding time, making a high cache hit rate critical.

 FlashMoE’s decoding stage also performs the cache replacement policy operations asynchronously, effectively overlapping them with expert loading latency to hide the overhead. On average, the FFN consumes 158µs per operation, whereas loading an expert from SSD takes roughly 3ms, which means FFN computation latency can be fully overlapped in expert loading. This design leverages loading latency in SSD-based inference, enabling not only lightweight cache policies like LRU and LFU, but more sophisticated cache policies with higher computational overhead like ML-Based Cache.

\subsection{ML-Based Cache Replacement Policy}

Since expert layer loading takes the most latency of the decoding stage, having a cache replacement with a higher hit rate algorithm is better for overall latency, despite its higher computational cost. Since the routing pattern of MoE models have both recency and frequency, we designed new cache policy with using feed-forward network, leveraging both LRU and LFU algorithm. Also, overall training process is shown in Fig. \ref{fig:training}, and will be discussed at the end of this section.

The following Algorithm. \ref{alg:mlcache} table illustrates the operating principle of the ML-based cache system proposed in FlashMoE. This cache system leverages LRU and LFU policies to complement the two approach. Specifically, it quantifies the recency and frequency information based on LRU and LFU for each expert and context, then applies a learning-based model to make more sophisticated eviction decisions.

The input to the cache policy model consists of recency and frequency scores for each expert, which are normalized for training before being fed into the model. The target data is generated based on the ideal, oracle\cite{belady}-based optimal experts to be evicted, with masked labels assigned to those experts.

\begin{algorithm}[t]
\caption{ML-Based Cache Replacement Policy}
\label{alg:mlcache}

\textbf{Input:} recency scores \(r_t\), frequency scores \(f_t\) \\
\textbf{Output:} Expert routing distance predictions and cache replacement decisions

\medskip

Normalize recency scores: \( r_{t,\mathrm{norm}} = \frac{1}{r_t} \) \\
Normalize frequency scores: \( f_{t,\mathrm{norm}} = \frac{f_t}{\max(f_t)} \) \\
Concatenate normalized scores: \( \text{score} = [r_{t,\mathrm{norm}}, f_{t,\mathrm{norm}}] \) \\
Compute distance prediction scores: \( \text{dist\_scores} = \text{ml\_cache}(\text{score}) \)

\medskip

\SetKwFunction{FMain}{CacheReplacement}
\SetKwProg{Fn}{Function}{:}{}
\Fn{\FMain{routed\_experts}}{
    \For{expert\_id \textbf{in} routed\_experts}{
        \If{cache.isHere(expert\_id) == False}{
            \If{cache.isFull() == True}{
                to\_evict = cache.MaxDist(dist\_scores)\;
                cache.evict(to\_evict)\;
            }
            cache.load(expert\_id)\;
        }
        expert\_w = cache.get(expert\_id)\;
        output += 
        routing\_w[expert] \(\times\) expert\_w(h\_state)\;
    }

    \Return output\;
}
\end{algorithm}

The trained cache prediction model, based on the input data and targets, is composed of a simple feed-forward network (FFN). During inference, an eviction score is calculated for each candidate expert, and the expert with the highest score is prioritized for eviction. Compared to traditional policies based on heuristics (LRU, LFU, LIFO), this approach enables more adaptive cache management and effectively reduces miss penalties through a precise replacement policy that reflects expert routing characteristics.

This formula explains how the recency and frequency scores used in the expert cache policy are calculated. The \textbf{recency score} represents how long it has been since the last access, indicating how many steps ago the expert was last accessed from the current time step. The \textbf{frequency score} measures how many times the expert has been routed during inference.

Let $r_t$ denote the recency score and $f_t$ the frequency score at time step $t$. These scores are updated from $r_{t-1}$ and $f_{t-1}$ as follows:

{\small
\begin{table}[t]
  \centering
  \caption{Training hyperparameters for FlashMoE}
  \label{table:hyperparam}
  \begin{tabular}{|l|l|}
    \hline
    \textbf{Hyperparameter} & \textbf{Value} \\
    \hline
    \hline
    Loss & MSELoss \\
    \hline
    Optimizer & AdamW \\
    \hline
    Learning Rate & 1e-3 \\
    \hline
    Learning Rate Decay & 1e-2 \\
    \hline
    \# Layer & 3 \\
    \hline
    Hidden size & 128 \\
    \hline
    Activation Func. & SiLU \\
    \hline
  \end{tabular}
\end{table}
}

\begin{itemize}
    \item If the expert is accessed at time $t$:
    \[
    r_t = 1,\quad f_t = f_{t-1} + 1
    \]
    \item If the expert is not accessed at time $t$:
    \[
    r_t = r_{t-1} + 1,\quad f_t = f_{t-1}
    \]
\end{itemize}

As shown in Generate Dataset in Fig.~\ref{fig:training}, the recency score is initialized such that when expert 4 is routed at $time\_step=0$, its recency score is set to 1, while unrouted experts remain at infinity. At $time\_step=1$, expert 3 is routed and thus marked as 1, while expert 4, not being routed, is incremented to 2. For the frequency score, expert 4 is marked as 1 at $time\_step=0$, and at $time\_step=1$, both experts 3 and 4 are marked as 1, corresponding to the count of routing events for each expert. The inputs to the FFN are normalized\cite{inputnorm}. Since $r_t$ and $f_t$ are not inherently normalized, we apply a normalization step. Because a smaller recency value should correspond to a higher routing probability, the recency score is mapped to the range $[0,1]$ using an inverse function:

\[
\text{Recency}_{\text{norm}} = \frac{1}{r_t}
\]

The frequency score is normalized by dividing by the maximum frequency observed so far, since larger frequency means higher chance of routing:

\[
\text{Frequency}_{\text{norm}} = \frac{f_t}{\max(f)}
\]

These two normalized scores are then concatenated and used as the input to the feed-forward network (FFN):

\[
\text{FFN Input} = \left[\text{Recency}_{\text{norm}}\,\, \|\,\, \text{Frequency}_{\text{norm}}\right]
\]

Fig. \ref{fig:training} and Table. \ref{table:hyperparam} each shows the data composition and hyperparameters used to train the ML-based cache policy. In this study, inference was performed on 512 samples from the TriviaQA\cite{TriviaQA} dataset, each up to 512 tokens, and the routing history during this process was extracted. Assuming Belady’s optimal cache replacement algorithm, masked target data indicating which experts should be evicted at each time step were generated. 

Preparing and training the ML-based cache involves four stages: routing trace extraction, feature construction, FFN training, and model evaluation. Assuming the model is OLMoE-1B-7B~\cite{olmoe}, while the exact time for dataset collection depends on batching, using a batch size of 32 on a single A100 GPU allows routing traces to be extracted in under 20 minutes. Feature construction dominates preprocessing but can be parallelized, and training the lightweight per-layer ML cache FFN($\simeq$ 113KB) is fast, allowing the entire pipeline to be completed within roughly 2 hours.

\begin{figure*}[t]
  \centering
  \begin{subfigure}{0.48\textwidth}
   \includegraphics[width=\linewidth]{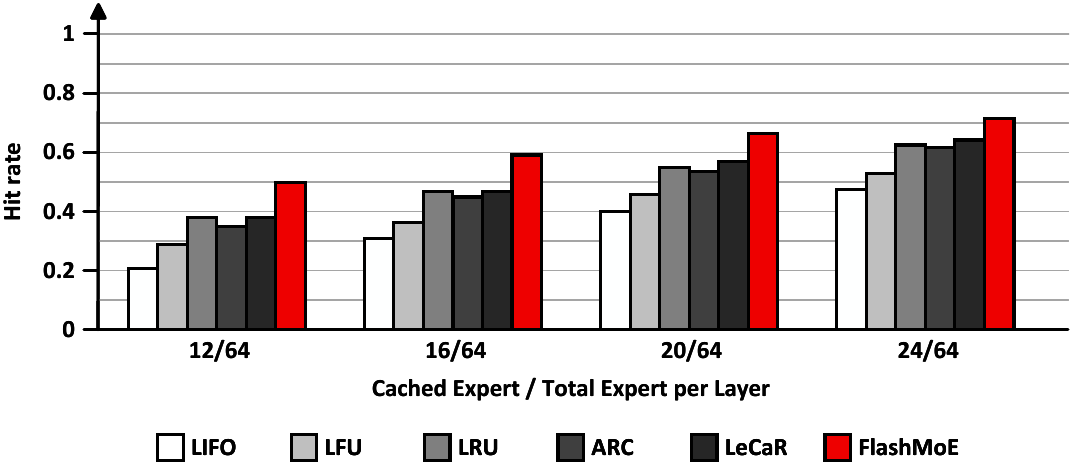}%
   \caption{}
  \end{subfigure}\hfill
  \begin{subfigure}{0.48\textwidth}
   \includegraphics[width=\linewidth]{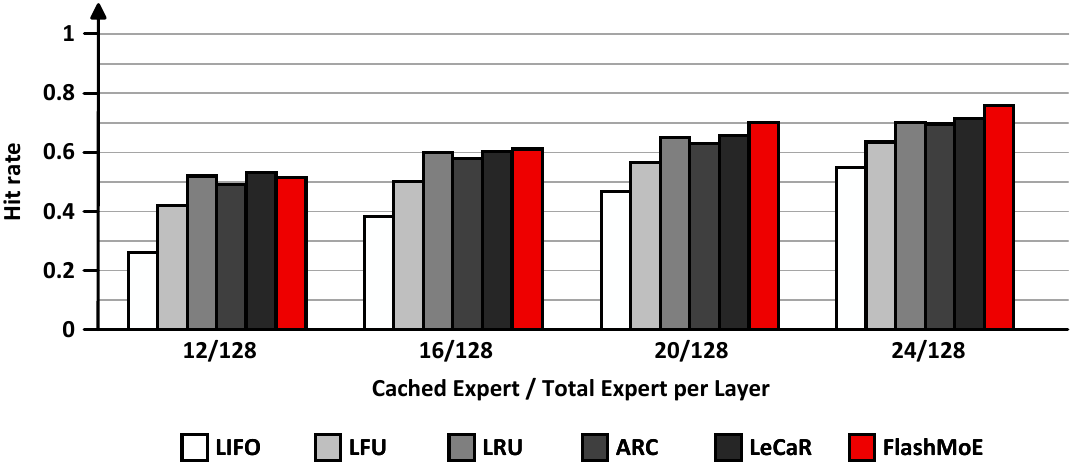}%
   \caption{}
  \end{subfigure}
  \caption{
    Hit rate by cache size \& replacement policy
    (a): OLMoE-1B-7B, (b): Qwen3-30B-A3B.
  }
  \label{fig:hitrate_combined}
\end{figure*}

\section{EVALUATION}

{\small
\begin{table}[!t]
  \centering
  \caption{Specifications of Desktop System used for FlashMoE.}
  \label{table:spec}
  \begin{tabular}{|l|l|}
    \hline
    \textbf{Component} & \textbf{Specification} \\
    \hline
    \hline
    CPU & AMD Ryzen 5 9600X (Zen 5, 6 cores, PCIe 5.0 support) \\
    \hline
    GPU & NVIDIA RTX 5070 Ti (16GB GDDR7, PCIe 5.0) \\
    \hline
    Motherboard & Gigabyte X870 AORUS EAGLE WiFi7 (PCIe 5.0) \\
    \hline
    SSD & SK hynix P51  NVMe SSD (PCIe 5.0, 7.4 GB/s read) \\
    \hline
    RAM & 16GB DDR5-6000MHz \(\times\) 4 (Dual Channel) \\
    \hline
    OS & Ubuntu 24.04 LTS \\
    \hline
    Environment & Python 3.11 with PyTorch 2.7, CUDA 12.9 \\
    \hline
  \end{tabular}
\end{table}
}

\subsection{Evaluation Environment}

For evaluation, we built a user-grade experimental environment equipped with high-speed interfaces, as summarized in Table.~\ref{table:spec}. To evaluate DRAM leverage, we deliberately left approximately 1GB of system DRAM available and locked the remaining memory to minimize interference. For model training, as described in Section 3, we utilized the TriviaQA\cite{TriviaQA} dataset. The question set used for decoding speed and cache hit-rate experiments consisted of a split of the dataset, which was not used in training. Training was performed using PyTorch with 200 epochs and early stopping.

We included Fiddler\cite{fiddler}, DAOP\cite{daop}, llama.cpp~\cite{llamacpp}, and HuggingFace Transformers~\cite{huggingface} as baselines. Fiddler and DAOP provide a configuration that initialize models with loading popular experts. To determine these experts, we extracted the routing history from the TriviaQA training dataset and identified the most frequently accessed experts as popular experts.

\begin{figure*}[t]
  \centering

  \begin{subfigure}{0.32\textwidth}
    \includegraphics[width=\linewidth]{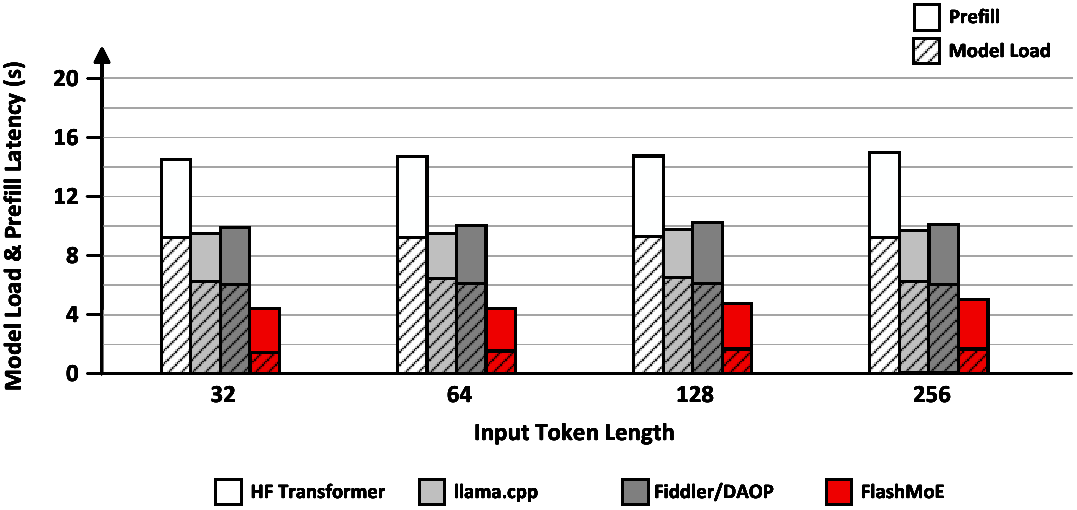}
    \caption{}
  \end{subfigure}\hfill
  \begin{subfigure}{0.32\textwidth}
    \includegraphics[width=\linewidth]{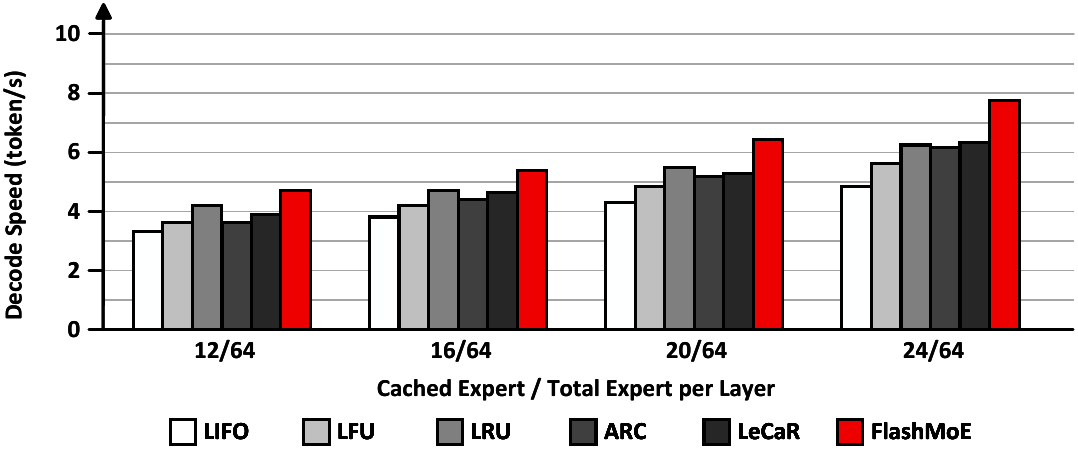}
    \caption{}
  \end{subfigure}\hfill
  \begin{subfigure}{0.32\textwidth}
    \includegraphics[width=\linewidth]{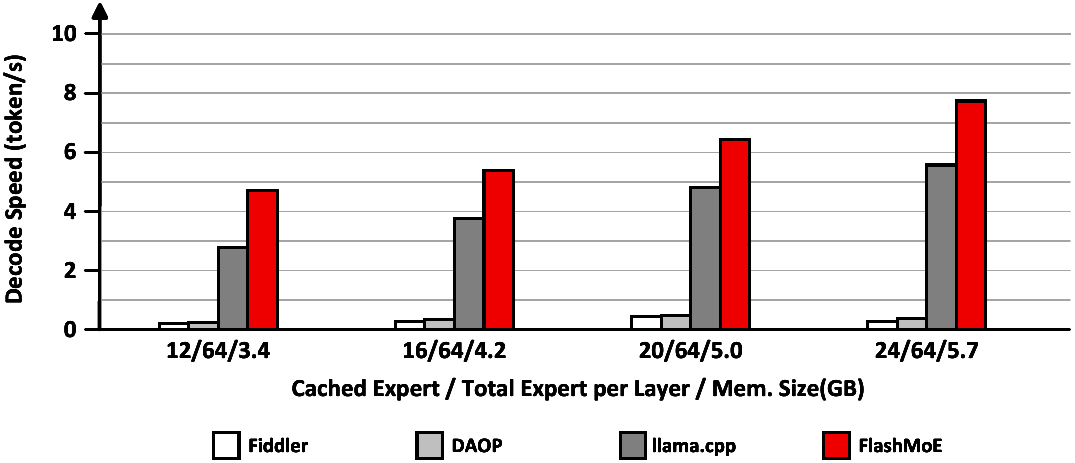}
    \caption{}
  \end{subfigure}

  \begin{subfigure}{0.32\textwidth}
    \includegraphics[width=\linewidth]{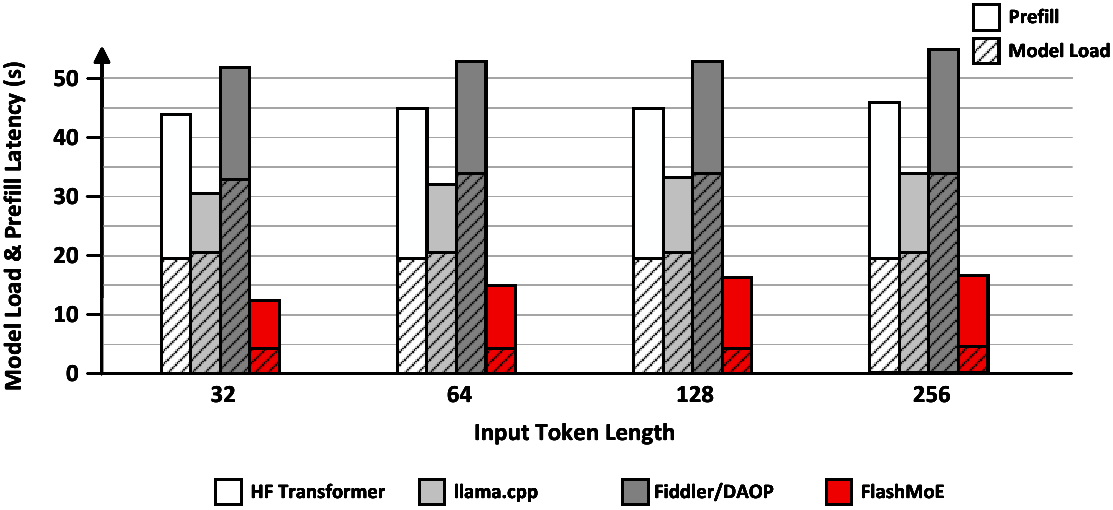}
    \caption{}
  \end{subfigure}\hfill
  \begin{subfigure}{0.32\textwidth}
    \includegraphics[width=\linewidth]{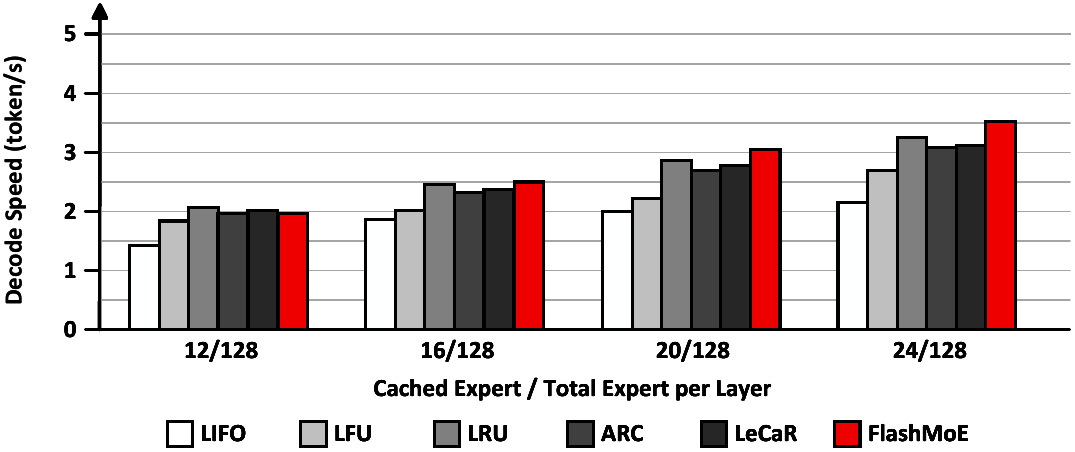}
    \caption{}
  \end{subfigure}\hfill
  \begin{subfigure}{0.32\textwidth}
    \includegraphics[width=\linewidth]{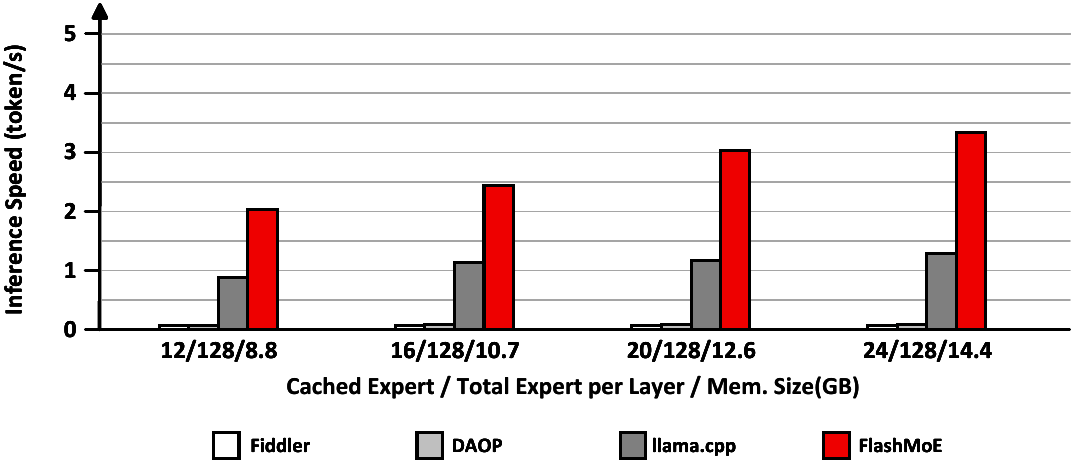}
    \caption{}
  \end{subfigure}

  \caption{
    (a,d) Model load \& prefill latency of the MoE inference system.  
    (b,e) Token generation speed under different cache replacement policies.  
    (c,f) Token generation speed across different inference systems. ((a,b,c): OLMoE-1B-7B, (d,e,f): Qwen3-30B-A3B)
  }
  \label{fig:all_moe_results}
\end{figure*}

\subsection{Initial Model Loading \& Prefill Latency}

Existing MoE inference systems\cite{fiddler, eliseev, daop} typically load the entire LLM model—either on the CPU or across CPU-GPUs—and only selectively activate expert layers during inference. In contrast, FlashMoE significantly reduces this overhead by separating expert and non-expert weights and loading only the non-expert weights at startup. As a result, FlashMoE achieves 4× faster initial loading than llama.cpp and 6.8× faster than Fiddler and DAOP.

We also analyzed prefill latency—the stage where experts are first loaded. As shown in Fig.~\ref{fig:all_moe_results}(a) and Fig.~\ref{fig:all_moe_results}(d), even when accounting for expert loading at prefill, FlashMoE maintains an overall latency advantage, yielding a total speedup of 2.5× over llama.cpp and 4.1× over Fiddler and DAOP.

A potential concern is that this advantage may diminish as the token length increases, due to the potential for more experts being routed. However, we found that the number of routed experts increases sublinearly with token length. About 47\% of experts were loaded when input token length equals to 32, and the proportion goes up to 58\%, 64\%, 67\%, while input length doubles up with 64, 128, 256. Thus, FlashMoE's performance advantage remains valid even for long input sequences.


\subsection{Cache Hit-Rate}
Fig. \ref{fig:hitrate_combined} illustrate cache hit rates of various replacement policies under different cache sizes for OLMoE-1B-7B and Qwen3-30B-A3B, respectively. Across all cache sizes, our proposed ML-based cache replacement policy consistently achieved the highest hit rate. Specifically, for OLMoE-1B-7B, it outperformed the widely adopted LRU by 21\% and LFU by 51\% each, which translates to a 22\% and 35\% reduction in I/O. In addition, when compared with adaptive and learning-based cache replacement policies such as ARC~\cite{arc} and LeCaR~\cite{lecar}, the proposed model achieved up to 28\% and 21\% higher hit rates, respectively. This confirms that the proposed model performs cache replacement more effectively not only than traditional heuristic-based policies but also than advanced cache management policies.

\subsection{Inference latency}

Under the same system conditions, we evaluated the impact of different cache policies on inference speed. For OLMoE-1B-7B (Fig. \ref{fig:all_moe_results}(b)) and Qwen3-30B-A3B (Fig. \ref{fig:all_moe_results}(e)), the ML-based cache consistently achieved the highest token throughput. Compared to the baseline LRU policy—the best-performing heuristic—the ML cache improved inference speed by approximately 22\% for OLMoE-1B-7B and 7\% for Qwen3-30B-A3B. This demonstrates that the proposed ML cache not only increases cache hit rates but also translates into tangible end-to-end latency benefits. While Fiddler and DAOP aim for DRAM-offloading-based inference, their CPU-based computation cannot be fully utilized due to memory bottlenecks. Furthermore, DAOP determine which experts to evict or load into VRAM by computing all possible combinations of expert layer placements. This approach may be feasible for models with a small number of experts (e.g., Mixtral-8x7B\cite{mixtral}), but it becomes impractical for more recent models with significantly more experts.

\section{Conclusion}
This paper proposes the \textbf{FlashMoE} system for efficient on-device inference of large-scale Mixture-of-Experts (MoE) models. FlashMoE enables large-scale model inference in memory-constrained environments by separately storing experts and non-experts and implementing SSD-based offloading. Additionally, it introduces an ML-based cache replacement policy. improving cache hit rates compared to traditional heuristic methods. By implementing FlashMoE in a real PC environment and conducting experiments on state-of-the-art LLM models such as Qwen3-30B-A3B and OLMoE-1B-7B, the study demonstrates significant inference performance within limited memory.


\bibliographystyle{ACM-Reference-Format}
\bibliography{sample-base}

\appendix

\end{document}